\documentclass[runningheads]{llncs}
\usepackage{graphicx}
\usepackage{amssymb,amsmath}
\usepackage{caption}
\usepackage{subcaption}
\usepackage{algorithm}
\usepackage{algpseudocode}
\usepackage{fancyvrb}
\usepackage[table, dvipsnames]{xcolor}
\usepackage{tcolorbox}
\usepackage{dsfont}
\usepackage{algorithm}
\usepackage{algpseudocode}
\usepackage{fancyvrb}
\usepackage{xcolor}
\usepackage{dsfont}
\usepackage{mathrsfs}
\usepackage{listings}
\usepackage{parskip}
\usepackage[shortlabels]{enumitem}
\usepackage{multirow}
\usepackage{multicol}
\usepackage{caption}
\usepackage{cite}

\usepackage[pagebackref,breaklinks,colorlinks]{hyperref}

\graphicspath{{images/}}

\newcommand{\E}{\mathbb{E}}
\newcommand{\kl}{D_\text{KL}}

\begin{document}
%
\title{Semi-Supervised Learning for\\ Deep Causal Generative Models}
%
%
\author{Yasin Ibrahim\inst{1} \and Hermione Warr\inst{1} \and Konstantinos Kamnitsas\inst{1,2,3}}

\authorrunning{Y. Ibrahim et al.}
%
\institute{Department of Engineering Science, University of Oxford, Oxford, UK\\
\email{\{firstname.lastname\}@eng.ox.ac.uk} \and
Department of Computing, Imperial College London, London, UK \and
School of Computer Science, University of Birmingham, Birmingham, UK
}
\maketitle              
\begin{abstract}
Developing models that are capable of answering
questions of the form ``How would \(x\) change if \(y\) had been \(z\)?'' is fundamental to advancing medical image analysis. Training causal generative models that address such counterfactual questions, though, currently requires that all relevant variables have been observed and that the corresponding labels are available in the training data. However, clinical data may not have complete records for all patients and state of the art causal generative models are unable to take full advantage of this. We thus develop, for the first time, a semi-supervised deep causal generative model that exploits the causal relationships between variables to maximise the use of all available data. We explore this in the setting where each sample is either fully labelled or fully unlabelled, as well as the more clinically realistic case of having different labels missing for each sample. We leverage techniques from causal inference to infer missing values and subsequently generate realistic counterfactuals, even for samples with incomplete labels. Code is available at: \url{https://github.com/yi249/ssl-causal}
  
  \keywords{Causal Inference \and Semi-Supervised \and Generative Models}
\end{abstract}
\section{Introduction}
\label{sec:intro}

The deployment of deep learning models to real-world applications faces a variety of challenges \cite{dlchallenge}, with many arguing that this is due to lack of causal considerations \cite{intro, Peters17}. A growing research area is the generation of counterfactuals (CFs), the manifestation of a sample in an alternative world where an upstream variable has been changed \cite{deepscm, neuralscm1, epi2}. Such techniques are particularly useful in medical image analysis, where models are often hampered by lack of diversity in training data \cite{causalitymatters}, so methods to generate realistic synthetic samples from underrepresented classes are critical \cite{nofairlunch}. Incorporating structural causal equations into a deep learning framework has been shown to provide a powerful tool for counterfactual generation \cite{deepscm}. These ideas were extended by the development of a hierarchical VAE structure for greater image fidelity \cite{hvae}. This method consists, however, of a separately trained generative model and structural causal model, represented by a directed acyclic graph (DAG), and hence, the two components are unable to leverage information from one another during training. Moreover, these methods rely on fully labelled samples, so are unable to use additional data where true values are unavailable for some (or all) variables of the causal graph.

Data with limited labels are ubiquitous, so semi-supervised methods are of particular interest. A common approach to semi-supervised learning is consistency regularisation under transformations of the input \cite{consistency1,consistency2,consistency3}. Alongside our generative model, we present this approach from a causal perspective and demonstrate how it fits naturally into our framework. Semi-supervised methods also have a causal motivation \cite{causalanticausal} due to the principle of independence of cause and mechanism (ICM) \cite{Peters17}, which suggests that possessing information on the effect (image) alone is beneficial for learning the joint distribution of cause (labels) and effect. In summary, we make the following contributions:

\begin{itemize}
    \item Introduce a semi-supervised deep causal generative model,
    \item Generate and evaluate counterfactuals with missing causal variables,
    \item Provide a causal perspective on the consistency regularisation technique for semi-supervised learning,
    \item Inspired by the ICM, investigate the performance of our method when parent variables are missing versus when child variables are missing.
\end{itemize} 

To illustrate this, we first use a semi-synthetic dataset based on Morpho-MNIST \cite{morphomnist} which allows us to explicitly measure performance given the known underlying causal relationships. We then assess our method on the MIMIC-CXR dataset \cite{mimic, physionet} to demonstrate its capabilities on real, more complex, medical data. 

\section{Background}
\label{sec:background}
A (Markovian) Structural Causal Model (SCM) is a 4-tuple \cite{neuralscm1}:
\[\mathcal{M} = \langle V,U,\mathcal{F}, P(U)\rangle\]
where, \(V = \{v_1, \dots, v_n\}\) is the set of endogenous variables of interest, \(U = \{u_1, \dots, u_n\}\) is the set of exogenous (noise) variables, \(P(U)\) is the prior distribution over them, and \(\mathcal{F} = \{f_1, \dots, f_n\}\) is a set of functions assigning values to the endogenous variables. Moreover, we assume that each endogenous variable, \(x_i\), is assigned deterministically by its direct causes, i.e. its parents \(\text{pa}_i \subseteq V\setminus \{v_i\}\), and the corresponding noise variable, \(u_i\) via the structural assignments,
\begin{equation}
    v_i := f_i(\text{pa}_i, u_i).
    \label{eq:func}
\end{equation}

This supersedes conventional Bayesian approaches as it allows for greater control by explicitly considering the structural relationships between variables. We achieve this through the do-operation \cite{foundation}, which makes assignments of the form \(\text{do}(x_i=a)\). This disconnects \(x_i\) from its parents, and we obtain an intervened distribution over the endogenous variables,
\[P(V|\text{do}(v_i=a)) = \prod_{j\neq i}p(v_j|pa_j)\cdot\mathds{1}_{\{v_i=a\}}\]
However, such interventions provide only population-level effect estimations \cite{foundation}. To narrow this down to unit-level and generate counterfactuals for individual samples, the following procedure is carried out\cite{Peters17}:
\begin{enumerate}
    \item \textbf{Abduction}: Use the data to update the prior probability on the exogenous noise \(p(U)\) to obtain \(p(U | V)\)
    \item \textbf{Action}: Perform an intervention \(\text{do}(V=A)\) to obtain a modified SCM, denoted by \(\mathcal{\Tilde{M}}_{\text{do}(V=A)}\).
    \item \textbf{Prediction}: Use \(\mathcal{\Tilde{M}}_{\text{do}(V=A)}\) to estimate the values of the desired variables.
\end{enumerate}

\section{Methodology}
\begin{figure}
    \centering
    \includegraphics[width=\textwidth]{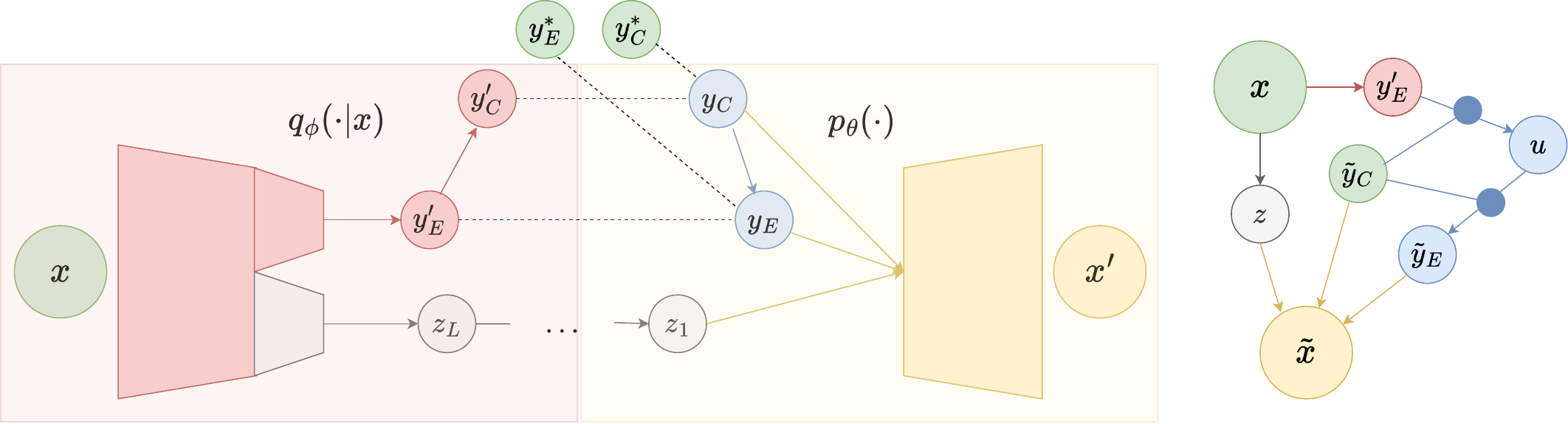}
    \caption{Model outline. Green: observed, Grey: latent, Red: predicted, Blue: causal generative, Yellow: decoding. (left) Training; we use the \(y\) predictions for decoding unless they are observed, (right) CF generation.}
    \label{fig:model}
\end{figure}
An overview of our method is shown in Fig.~\ref{fig:model}. Herein, the endogenous variables consist of image, \(x\), and variables, \(y\); denoted by \(y^*\) when observed and by \(y'\) when predicted. Latent variables \(z=z_{1:L}\) make up part of the exogenous noise for \(x\), modelled using a hierarchical latent structure, following \cite{hvae, ladder}. Our model extends this structure with a predictive part that infers the endogenous variables, \(y\), enabling counterfactual generation in the case of missing labels. For clarity, we limit derivations to the case with a single cause variable, \(y_C\), and effect variable, \(y_E\), but this can be extended to any finite number of variables with any causal structure. The ELBO loss for \textbf{fully labelled samples} drawn from \(\mathcal{D}_L\) is \(\mathcal{S}(x,y)\):
\begin{align*}
    &\log p_\theta(x) \; \geq \; \E_{q_\phi(z|x,y_E,y_C)}\left[\log \frac{p_\theta(x|z,y_E,y_C)p_\theta(z)p_\theta(y_E|y_C)p_\theta(y_C)}{q_\phi(z|x,y_E,y_C)}\right] \\
    &\Rightarrow \mathcal{S}(x,y) := \; -\mathcal{L}(x,y) - \log p_\theta(y_E|y_C) - \log p_\theta(y_C)
\end{align*}
where \(\mathcal{L}(x,y)\) is the ELBO for a conditional VAE and \(p_\theta(\cdot)\) are (Gaussian) priors. For the \textbf{unlabelled samples}, drawn from \(\mathcal{D}_U\), we minimise the loss:
\begin{align}
\begin{split}
    \mathcal{U}(x) := -\E_{q_\phi(y_E|x)}\big[&\E_{q_\phi(y_C|x,y_E)}(\mathcal{L}(x,y)
    - \kl\{q_\phi(y_C|x,y_E)||p_\theta(y_C)\})\big]\\
    + &\kl\{q_\phi(y_E|x)||p_\theta(y_E|y_C)\}.
\end{split}
\label{eq:bothunlab}
\end{align}
Here, we predict the labels using \(y' \sim q_\phi(y|x)\) and regularise them via the KL-divergence with their respective prior \(p_\phi(y)\). When \textbf{only cause \(\mathbf{y_C}\) is labelled}, for samples \((x,y_C) \in \mathcal{D}_C\), we minimise the loss:
\begin{equation}
\mathcal{C}(x,y_C) := -\E_{q_\phi(y_E|x)}[\mathcal{L}(x,y)] - \log p_\theta(y_C) + \kl\{q_\phi(y_E|x)||p_\theta(y_E|y_C)\}.
\label{eq:cause}
\end{equation}
When \textbf{only effect \(\mathbf{y_E}\) is labelled}, we minimise the loss:
\begin{equation*}
\mathcal{E}({x,y_E}) := -\E_{q_\phi(y_C|x,y_E)}[\mathcal{L}(x,y) + \log p_\theta(y_E|y_C)]+ \kl\{q_\phi(y_C|x,y_E)||p_\theta(y_C)\},
\end{equation*}
for samples \((x,y_E) \in \mathcal{D}_E\). In the case of discrete variables, when the true labels, \(y*\), are not provided, we supplement these losses by inversely weighting samples with missing labels by the entropy of the labels predicted by the encoder, \(H_\phi(y'|x)\). For example, when \(y_E\) is missing, we multiply the expectation in \eqref{eq:cause} by \(1-H_\phi(y'_E|x)\). We use entropy here as an indicator for predictive uncertainty, to inform us how much to `trust' the predicted label.

Under the current construction, the parent predictors, \(q_\phi(y_E|x)\) and \(q_\phi(y_C|x,y_E)\), are only trained when the parent variables are unobserved. To ensure the model is able to learn a good representation of \(y\), we include additional classification terms in the supervised loss \cite{semisup}, giving the \textbf{total loss} to train our model:
\begin{align}
\begin{split}
\mathcal{T}(x,y) := &\sum_{(x,y)\in \mathcal{D}_L} \mathcal{S}(x,y) + \sum_{x \in \mathcal{D}_U}\mathcal{U}(x) + \hspace{-1em}\sum_{(x,y_C)\in \mathcal{D}_C}\hspace{-0.5em} \mathcal{C}(x,y_C)\\ 
+ &\sum_{(x,y_E)\in \mathcal{D}_E} \hspace{-0.5em}\mathcal{E}({x,y_E}) - \E_{(x,y) \in \mathcal{D}_L} \left[\log q_\phi(y_i|y_{<i},x)\right]
\end{split}
\end{align}

In the last term, labeled variables \(y_i\) are placed in a topological ordering \cite{topolog} starting with the root nodes. Thus for all \(i\), the ancestors of \(y_i\) are a subset of \(y_{<i}\) and its descendants are a subset of \(y_{>i}\). In \eqref{eq:bothunlab}, we see that in the unlabelled case, we require expectations over the labels. For a single discrete label \(y\), this can be achieved by summing over all possible values of \(y\) \cite{semisup},
\[\E_{q_\phi(y|x)}[f(y,\cdot)] = \sum_y q_\phi(y|x)\cdot f(y,\cdot).\]
However, this quickly becomes computationally expensive as the number of variables or classes grows, and is intractable for continuous variables. We avoid using a Monte-Carlo sampler, where taking more samples leads to similar computational costs. Instead, we propose lowering this variance by beginning training using only the labelled data. By doing so, predictors \(q_\phi(\cdot|x)\) reach a sufficiently high accuracy before being used to estimate missing labels for the rest of the data.

\subsubsection{Counterfactual Regularisation}
To further improve our model, we turn to a causal treatment of consistency regularisation \cite{consistency1}. For this, we restrict perturbations of the input image to interventions on the DAG governing the causal variables. For example, for input image \(x\) with variables \((y_C, y_E)\), we alter effect variable \(y_E\) via \(\text{do}(y_E = \Tilde{e})\), to obtain new image \(\Tilde{x}\) with causal variables \((\Tilde{y}_C, \Tilde{y}_E)\). If the DAG is obeyed, the cause variable should remain invariant to this perturbation and we can thus impose a loss that penalises divergence between \(y_C\) and \(\Tilde{y}_C\). In the context of our model, we predict \(\Tilde{y}_C \sim q_\phi(\cdot|\Tilde{x})\) and minimise \(D(y_C, \Tilde{y}_C)\), where \(D(\cdot, \cdot)\) is an appropriate distance metric. If \(y_C\) is unknown, we predict its value using \(y_C \sim q_\phi(\cdot|x)\). Suppose instead we alter the cause variable; in this case, the effect should change given the DAG. As such, we predict \(\Tilde{{y}_E} \sim q_\phi(\cdot|\Tilde{x})\) and then compute the counterfactual of \(y_E\) under the proposed intervention on \(y_C\). When \(y_E\) is unlabelled, we first predict it using \(y_E \sim q_\phi(\cdot|x)\). This causal interpretation improves the robustness not only of the generative component of the model, but also the causal inference elements, \(p_\theta(y_i|y_{<i})\).
\subsubsection{Counterfactual Generation}
Once the generative model is trained, we use the predictive component \(p_\theta(y_i|y_{<i})\) to generate counterfactuals, \(\Tilde{x}\). For abduction we require the structural assignments \eqref{eq:func} to be invertible in \(u\), so we encode \(p_\theta(y_i|y_{<i})\) as an invertible
\(g_{y_{<i}}(u), \; u \sim p(u)\), parameterised by the causal parents \(y_{<i}\) \cite{deepscm}. Counterfactual generation can then be expressed as \(\Tilde{y}_i = g_{\Tilde{y}_{<i}}\left(g_{y_{<i}}^{-1}(y_i)\right)\) so that each counterfactual value can be calculated in sequence. If a label \(y_i\) is missing, we use our trained predictor \(q_\phi(y_i|x,y_{>i})\) to impute it, e.g Fig.~\ref{fig:model} (right) for when \(y_E\) is unobserved and we intervene on \(y_C\).
\section{Experiments}
\begin{figure}
\centering
\includegraphics[width=0.9\textwidth]{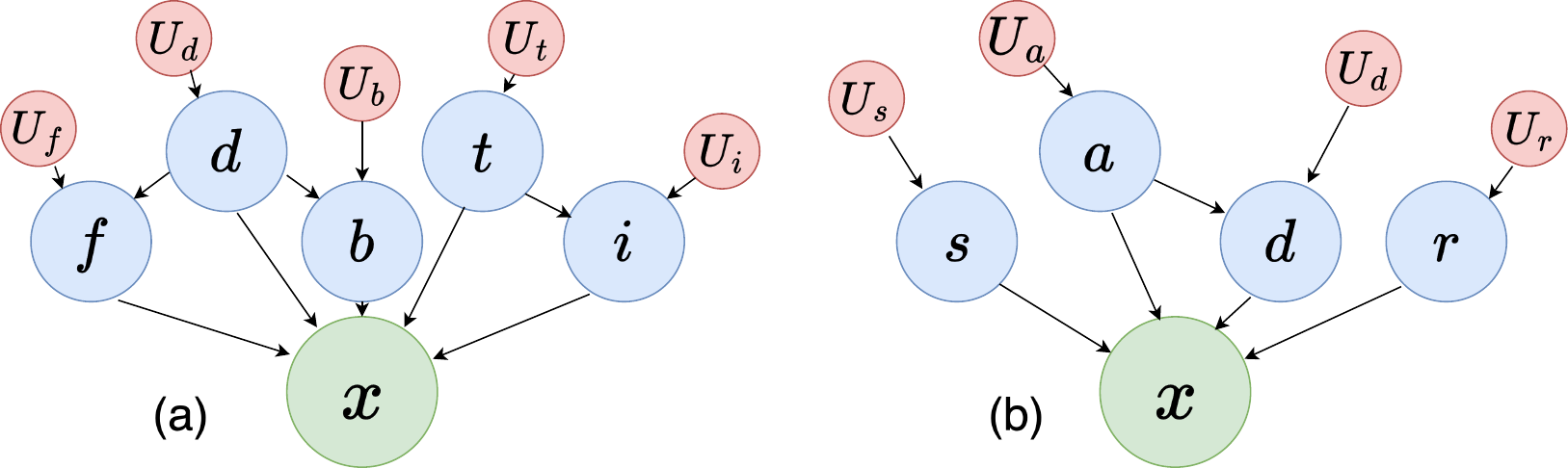}
\caption{(a) DAG for Colour MorphoMNIST, \(d\): digit, \(f\): foreground (digit) color, \(b\): background color, \(t\): thickness, \(i\): intensity. (b) DAG for MIMIC-CXR, \(s\): sex, \(a\): age, \(d\): disease status, \(r\): race. \(U\): respective exogenous noise variables.}
\label{fig:dags}
\end{figure}
\begin{figure}
		\centering
		\begin{subfigure}[b]{0.48\textwidth}
			\centering
			\includegraphics[width=0.8\textwidth]{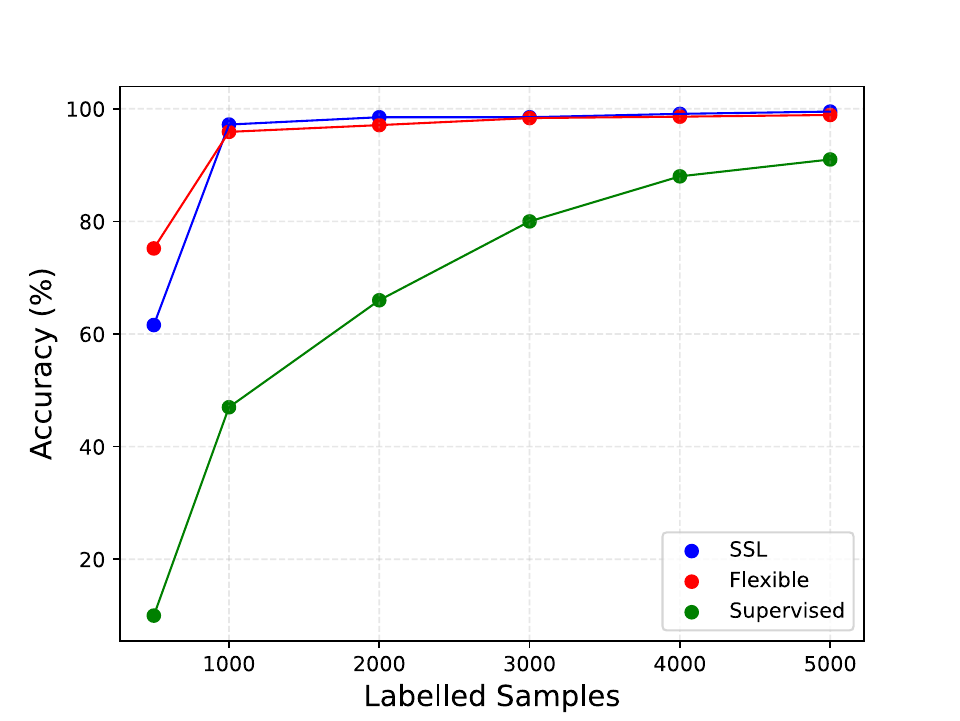}
			\caption[]%
			{{\small Supervised vs SSL vs Flexible.}}    
			\label{fig:dodigita}
		\end{subfigure}
		\hfill
		\begin{subfigure}[b]{0.48\textwidth}  
			\centering 
			\includegraphics[width=0.8\textwidth]{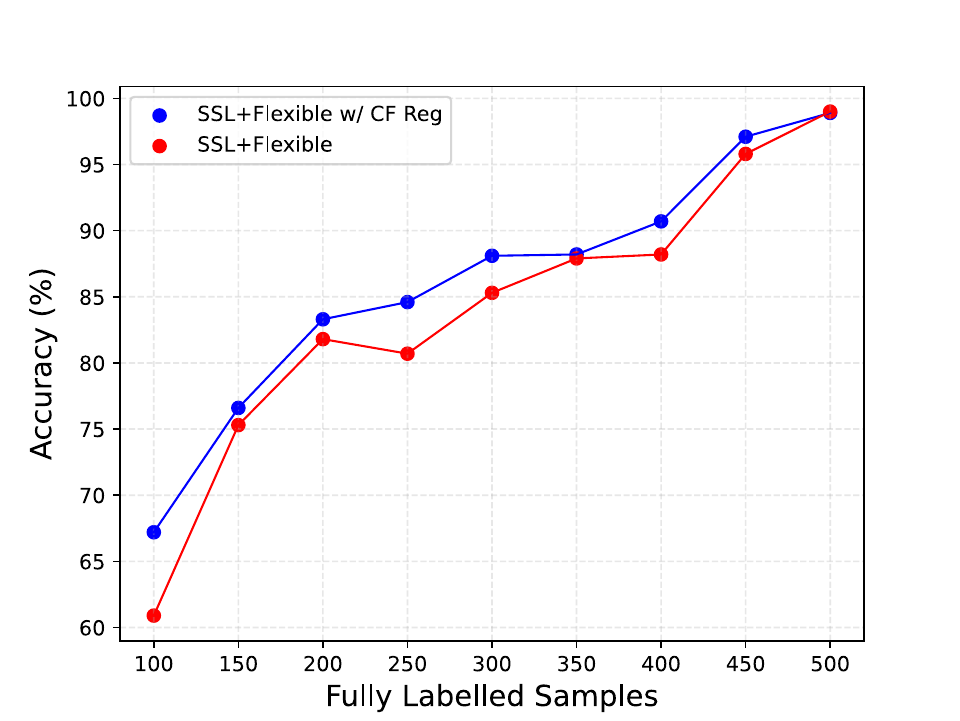}
			\caption[]%
			{{\small SSL+Flexible for very few labels.}}     
			\label{fig:dodigitb}
		\end{subfigure}
		\caption[]%
			{Colour Morpho-MNIST: Accuracy of \(\text{do}(d=k)\) on random test images for uniformly random \(k \in \{0,\dots,9\}\setminus d\) where \(d\) is the digit of the original image. For SSL, the \(x\)-axis represents to the number of fully labelled samples; for Flexible it represents the number of labels for each variable across all the samples. For SSL+Flexible we use 600 randomly allocated labels for each variable in addition to the number of fully labelled samples denoted by the \(x\)-axis.}
		\label{fig:dodigit}
\end{figure}
\subsection{Causal Analysis on Colour Morpho-MNIST}
\label{sec:mnist}
True causal generative processes for medical imaging data are unknown. Therefore, to evaluate the causal aspects of our method in depth, we first use data adapted from Morpho-MNIST (60k training, 10k test samples) \cite{morphomnist}, considering the thickness (\(t\)) and intensity (\(i\)) of each digit. We increase the complexity of the causal structure by colouring the foreground (\(f\)) and background (\(b\)), with each colour drawn from a distribution conditional on the digit (\(d\)), as in Fig.~\ref{fig:dags}\textcolor{red}{a}.
\subsubsection{Counterfactual Effectiveness}
As baseline, we use the state of the art supervised method for counterfactual image generation \cite{hvae} trained only on the labelled samples of each experiment. We compare this against our method for a labelled-unlabelled split (SSL in figures) and for labels missing randomly for each variable (Flexible). We measure the effectiveness \cite{axiomatic} of our counterfactual generations by abducting and intervening on test images with random interventions before using classifiers or regressors, \(q(\cdot|x)\), trained independently on uncorrelated data, to measure how well the desired change has been captured.
\begin{center}
\captionof{table}{Colour Morpho-MNIST: Log likelihoods \((\uparrow)\) of the child variables. Colour log likelihoods \(\in(-\infty, \: 2.239] \), intensity log likelihoods \(\in(-\infty, \: -1.336]\). }
\scalebox{0.8}{
\begin{tabular}{c|c|c|c|c|c|c|c|c} 
 Model & Labelled & \(\; \text{MAE} \; (\downarrow)\) &\(\; p_\theta(f) \;\) & \(\; p_\theta(b) \;\) & \(\; p_\theta(i,t) \;\) &\(\; q(f|\Tilde{x}) \;\) & \(\; q(b|\Tilde{x}) \;\) & \(\; q(i,t|\Tilde{x}) \;\)\\
 \hline
 \multirow{3}{*}{Supervised} & 1000 & 3.91 &-1.46 & -1.49 & -38.21 & -1.01 & -1.17 & -32.98\\
     & 5000 & 3.84 &-1.31 & -1.38 & -21.38 & -0.63 & -0.93 & -28.44 \\
     & 60,000 & 3.75 &1.20 & 1.24 & -5.55 & 1.17 & 1.18 & -14.42\\                           
 \hline
 \multirow{2}{*}{SSL} & 1000 & 3.85 & 1.05 & 1.10 & -14.26 & 0.81 & 1.08 & -26.02\\
  & 5000 & 3.83 &1.10 & 1.12 & -7.40 & 1.01 & 1.19 & -22.39\\
 \hline
 \multirow{2}{*}{Flexible} & 1000 & 3.86 &1.11 & 1.13 & -17.93 & 0.77 & 1.15 & -28.20\\
   & 5000 & 3.84 &1.14 & 1.16 & -13.98 & 1.07 & 1.10 & -19.56\\
 \hline
\end{tabular}
}
\label{tab:dos}
\end{center}
Fig.~\ref{fig:dodigita} highlights the improvement by our method over the purely supervised approach. Even when only 1000 samples \((\sim 1.67\%)\) are labelled, we achieve near perfect effectiveness for changed digit counterfactuals. This holds both when the data is split into distinct labelled-unlabelled sets and when these labels are missing randomly. Moreover, in Fig.~\ref{fig:dodigitb}, counterfactual  regularisation improves performance for very low labelled sample sizes by an average of \(\sim 2.2\%\). Table \ref{tab:dos} demonstrates how the causal relationships are learned significantly better using our method, with regards to both the distributions inferred from the DAG, \(p_\theta(\cdot)\), and the manifestations of these changes in the counterfactual image, \(q(\cdot|\Tilde{x})\).

\subsubsection{Independence of Cause and Mechanism}
\label{sec:causeandmech}
Inspired by insights on the ICM \cite{causalanticausal}, we analyse how our method performs in the specific cases of the cause variable missing and the effect present, and vice-versa, by varying the number of thickness and intensity labels while keeping the others. Table \ref{tab:causheandmech} (left) shows that the settings with greater proportions of intensity (effect) labels tend to produce better joint distributions, supporting the ICM hypothesis \cite{causalanticausal}. This is significant for domains with limited labelled data such as healthcare, as it suggests that, given an identified cause-effect relationship and limited capability to obtain labels, focusing on labelling the effect should provide improved performance.
\begin{figure}
  \captionof{table}{(left) Colour Morpho-MNIST: Cause and mechanism experiment. (right) MIMIC-CXR: For each intervention $do(\cdot)$, the 3 rows correspond to training with 10\%, 20\%, 30\% of variables labelled, all using CF regularisation. Semi-supervision in both settings (SSL, Flexible) outperforms pure supervision (Sup.).
  }  
  \begin{minipage}[]{.48\textwidth}
    \centering
    \scalebox{0.72}{
    \begin{tabular}{c|c|c|c|c|c} 
 \multirow{3}{*}{\begin{tabular}{@{}c@{}}\(i\) \\ labels \end{tabular}} & \multirow{3}{*}{\begin{tabular}{@{}c@{}}\(t\) \\ labels \end{tabular}} & \multicolumn{2}{c|}{\(\text{do}(t)\)} & \multicolumn{2}{c}{\(\text{do}(i)\)} \\ 
 \cline{3-6}
 \rule{0pt}{4ex} & & \(p_\theta(i,\!t) \; (\uparrow)\) & \(q(i,\!t|\Tilde{x}) \; (\uparrow)\) & \(|t\! -\! \Tilde{t}| \;(\downarrow)\) & \(|i\! -\! \Tilde{i}| \; (\downarrow)\)\\[1.5ex]
 \hline
  300 & 2700 & -21.84 & -29.47 & \textbf{0.089} & 0.293 \\
  600 & 2400 & -18.89& -26.17 & 0.102 & 0.203\\
  1200 & 1800 & -18.01 & -24.78 & 0.099 & 0.221\\
  1500 & 1500 & -19.40 & -25.43 & 0.132 & 0.172 \\
  1800 & 1200 & -16.27 & -22.13 & 0.120 & 0.161\\
  2400 & 600 & -14.62 & -19.32 & 0.146 & 0.093\\
  2700 & 300 & \bf{-14.15} & \bf{-17.56} & 0.152 & \bf{0.088}\\
 \hline
\end{tabular}}
    \label{tab:causheandmech}
  \end{minipage}\hfill
  \begin{minipage}[]{.48\textwidth}
    \centering
    \scalebox{0.53}{
    \begin{tabular}{c|c|ccc|ccc|ccc|ccc}
  \multicolumn{2}{c|}{} & \multicolumn{3}{c|}{Disease \((\uparrow)\)} & \multicolumn{3}{|c|}{Age \((\downarrow)\)} & \multicolumn{3}{|c|}{Sex \((\uparrow)\)} & \multicolumn{3}{|c}{Race \((\uparrow)\)}\\
  \cline{3-14}
  \multicolumn{2}{c|}{} & Sup. & SSL & Flex. & Sup. & SSL & Flex. & Sup. & SSL & Flex. & Sup. & SSL & Flex.\\
 \hline
  \multirow{3}{*}{do\((d)\)} & 10\% & 0.55 & 0.70 & \textbf{0.71} & 15.62 & 9.12 & 9.19 & 0.95 & 0.95 & 0.94 & 0.71 & 0.71 & 0.68\\
  & 20\% & 0.57 & \textbf{0.78} & 0.77 & 14.21 & 8.65 & 8.52 & 0.94 & 0.99 & 0.99 & 0.74 & 0.77 & 0.77\\
  & 30\% & 0.68 & 0.83 & \textbf{0.84} & 12.95 & 8.53 & 7.82 & 0.98 & 1.00 & 0.95 & 0.77 & 0.82 & 0.81\\
  \hline
  \multirow{3}{*}{do\((a)\)} & 10\% & 0.87 & 0.91 & 0.93 & 15.01 & \textbf{13.38} & 13.75 & 0.90 & 0.96 & 0.90 & 0.72 & 0.81 & 0.77\\
  & 20\% & 0.87 & 0.96 & 0.96 & 15.40 & \textbf{12.57} & 13.21 & 0.94 & 1.00 & 0.99 & 0.71 & 0.83 & 0.80\\
  & 30\% & 0.90 & 0.96 & 0.95 & 14.26 & \textbf{12.07} & 12.15 & 0.98 & 1.00 & 1.00 & 0.78 & 0.85 & 0.84\\
  \hline
  \multirow{3}{*}{do\((s)\)} & 10\% & 0.84 & 0.91 & 0.91 & 14.15 & 9.31 & 9.31 & 0.69 & \textbf{0.97} & 0.90 & 0.69 & 0.74 & 0.77\\
  & 20\% & 0.86 & 0.97 & 0.96 & 13.57 & 8.25 & 7.87 & 0.73 & \textbf{1.00} & 0.99 & 0.69 & 0.80 & 0.83\\
  & 30\% & 0.89 & 0.97 & 0.98 & 12.92 & 7.99 & 7.95 & 0.78 & \textbf{0.99} & 0.99 & 0.77 & 0.80 & 0.82\\
  \hline
  \multirow{3}{*}{do\((r)\)} & 10\% & 0.84 & 0.93 & 0.95 & 15.08 & 9.76 & 9.73 & 0.96 & 0.98 & 0.95 & 0.46 & \textbf{0.53} & 0.52\\
  & 20\% & 0.88 & 0.95 & 0.95 & 14.27 & 7.87 & 7.79 & 0.95 & 1.00 & 0.99 & 0.50 & \textbf{0.57} & 0.56\\
  & 30\% & 0.93 & 0.96 & 0.97 & 14.11 & 7.37 & 7.61 & 0.98 & 1.00 & 1.00 & 0.55 & 0.62 & \textbf{0.63}\\
  \hline
\end{tabular}}

\label{tab:tab2}
  \end{minipage}
\end{figure}
\subsection{Counterfactuals for Medical Imaging Data}
To evaluate our method on medical imaging data, we apply it to the MIMIC-CXR dataset (50k training, 20k test samples) \cite{mimic, physionet}. We assume the casual structure used in \cite{hvae} with variables disease (\(d\)), age (\(a\)), sex (\(s\)), race (\(r\)), with the only non-independence being that \(a\) causes \(d\) (Fig.~\ref{fig:dags}\textcolor{red}{b}). For disease, we use the presence of pleural effusion as a binary variable and we train models using 10\%, 20\%, 30\%, 40\% and 50\% of the total labels for each of the three models (Supervised, SSL, Flexible). As the causal structure is simpler, we measure performance, over 3 seeds, by intervening on each variable separately before measuring the ROCAUC for the discrete variables and the MAE for age.\\
\begin{figure}
  \begin{minipage}[]{.48\textwidth}
    \centering
    \includegraphics[width=0.88\textwidth]{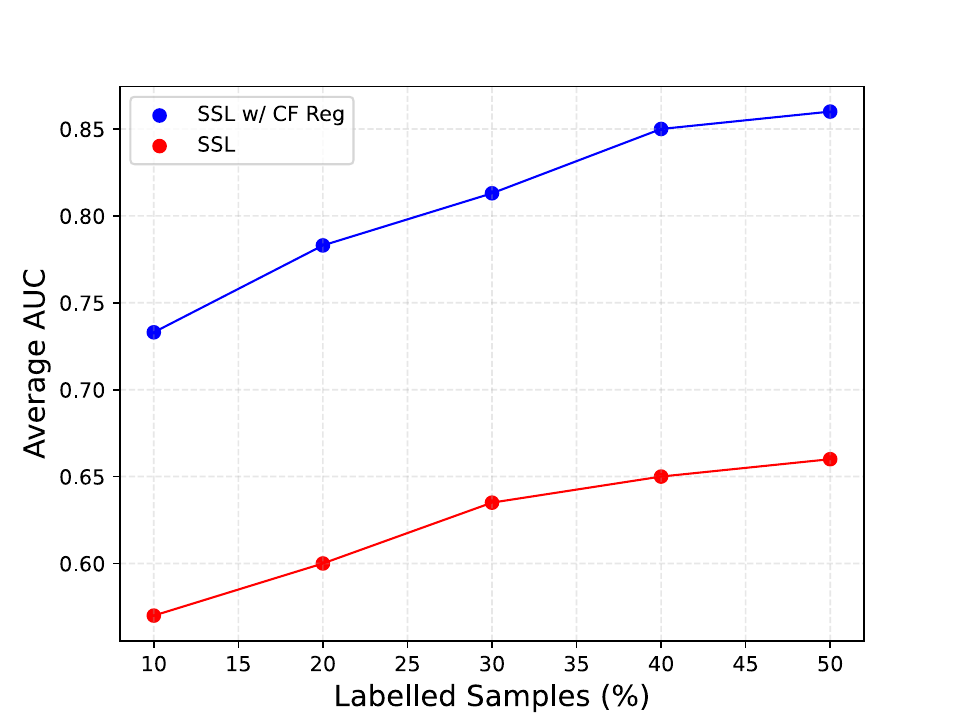}
    
  \end{minipage}\hfill
    \begin{minipage}[]{.48\textwidth}
    \centering
    \includegraphics[width=0.85\textwidth]{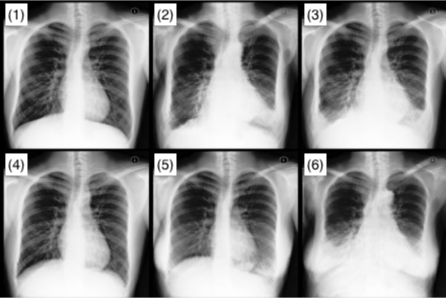}
  \end{minipage}
  \captionof{figure}{(a) CF Regularisation on MIMIC-CXR. (b) MIMIC-CXR CFs from model trained on 40\% labels. From top-left: (1) original: white, healthy, 20-year-old male, (2) do(\(\text{age}\!=\!90\)), (3) do(diseased), (4) do(asian), (5) do(female), (6) do(all).}
  \label{fig:mimic}
\end{figure}

From the cells on the diagonal of Table \ref{tab:tab2} (right), we see that our method tends to improve upon the supervised approach with regards to implementing interventions. The other cells are essentially a measure of reconstruction quality, since they involve evaluating variables that have not been intervened on. As such, the closeness of these values for the various models suggests that the achieved counterfactual generation gains are primarily due to differences in the causal inference component. This indicates that it would be fruitful to focus future efforts on improving this section of the model. This holds for both SSL and Flexible, demonstrating that practitioners implementing our approach need not prioritise achieving full labelling for any given sample over collecting as many labels as possible, bolstering the usability of the model. Fig.~\ref{fig:mimic}\textcolor{red}{a} demonstrates the increased interventional accuracy provided by CF regularisation. Moreover, as shown in Figure \ref{fig:mimic}\textcolor{red}{b}, our model is able to exhibit clear visual changes for the various CFs, indicating the numerical results are not due to minute changes undetectable to the human eye \cite{readingrace}. To build upon this, an avenue of future research would be to use this approach to generate additional training data for underrepresented populations in medical datasets and evaluate how this aids downstream tasks.

\section{Conclusion}
This study introduces a semi supervised deep causal generative model to enable training on causal data with missing labels in medical imaging. Experiments on a coloured Morpho-MNIST dataset, where the whole generative process is known, along with experiments on real clinical data from MIMIC-CXR, demonstrate that our approach uses unlabelled and partially labelled data effectively and improves over the state of the art fully supervised causal generative models. The key practical contribution of this work is that it enables training causal models on clinical databases where patient data may have missing labels, which previous models could not use, relaxing one of the main requirements for training a causal model. A limitation of our work is that we assume the DAG structure is known a priori. Hence, if this is misspecified, there are no guarantees on the correctness of the generated counterfactuals. A possible next step could thus be to explore cases with limited information on the DAG structure of the causal variables. 

\begin{credits}
\subsubsection{\ackname} Yasin Ibrahim and Hermione Warr are supported by the EPSRC Centre for Doctoral Training in Health Data Science (EP/S02428X/1). The authors also acknowledge the use of the University of Oxford Advanced Research Computing (ARC) facility in carrying out this work (http://dx.doi.org/10.5281/zenodo.22558).

\subsubsection{\discintname}
The authors have no competing interests to declare that are relevant to the content of this article.
\end{credits}

\bibliographystyle{splncs04}
\bibliography{bib}
\end{document}